\documentclass{article}

\usepackage[numbers,compress]{natbib}

\usepackage[preprint]{neurips_fitml_2024}

\usepackage{bm} 
\usepackage[utf8]{inputenc} 
\usepackage[T1]{fontenc}    
\usepackage{url}            
\usepackage{booktabs}       
\usepackage{amsfonts}       
\usepackage{nicefrac}       
\usepackage{color}
\usepackage[dvipsnames,table,xcdraw]{xcolor}

\usepackage[utf8]{inputenc} 
\usepackage[T1]{fontenc}    
\usepackage{hyperref}       
\usepackage{url}            
\usepackage{booktabs}       
\usepackage{amsfonts}       
\usepackage{nicefrac}       
\usepackage{microtype}      
\usepackage{algpseudocode}
\usepackage{amsmath, amssymb, amsthm}

\usepackage{float, fullpage, graphicx, multirow,parskip, subcaption, setspace}
\usepackage{comment}
\usepackage{url}
\usepackage{enumitem}
\usepackage{tikz}
\usepackage{bm, bbm}

\usepackage{rotating}
\usepackage{hyperref}
\usepackage{natbib}
\usepackage{placeins}
\usepackage[ruled,vlined]{algorithm2e}
\SetKwInput{kwInit}{Init}

\usepackage{thm-restate}

\declaretheorem[name=Proposition]{reproposition}
\declaretheorem[name=Lemma]{relemma}

\declaretheorem[name=Remark]{reremark}

\title{OCMDP: Observation-Constrained Markov Decision Process}

\author{
    \textbf{Taiyi Wang$^{1*\dag}$, 
    Jianheng Liu$^4$\thanks{Equal Contribution.} \ , 
    Bryan Lee$^1$, 
    Zhihao Wu$^3$,}
    \textbf{Yu Wu$^2$}\thanks{Corresponding Email: Taiyi.Wang@powersense.tech, yw573@cam.ac.uk} \\
    $^1$Powersense Technology Limited, $^2$University of Cambridge \\
    $^3$University of Edinburgh \\
    $^4$University College London \\
}

\begin{document}

\maketitle

\begin{abstract}
In many practical applications, decision-making processes must balance the costs of acquiring information with the benefits it provides. Traditional control systems often assume full observability, an unrealistic assumption when observations are expensive. We tackle the challenge of simultaneously learning observation and control strategies in such cost-sensitive environments by introducing the Observation-Constrained Markov Decision Process (OCMDP), where the policy influences the observability of the true state. To manage the complexity arising from the combined observation and control actions, we develop an iterative, model-free deep reinforcement learning algorithm that separates the sensing and control components of the policy. This decomposition enables efficient learning in the expanded action space by focusing on when and what to observe, as well as determining optimal control actions, without requiring knowledge of the environment's dynamics. We validate our approach on a simulated diagnostic task and a realistic healthcare environment using \textit{HeartPole}. Given both scenarios, the experimental results demonstrate that our model achieves a substantial reduction in observation costs on average, significantly outperforming baseline methods by a notable margin in efficiency.
\end{abstract}




\section{Introduction}

In traditional control systems, it is often assumed that all necessary information is readily available, which is seldom the case in practical scenarios~\cite{buarbulescu2024learned,wang2024distrl,wang2024ia2,wangenhancing,wang2021solving}. The need to actively decide which observations to make adds a layer of complexity to the decision-making process, as it requires balancing the benefits of additional information against the costs of acquiring it~\cite{haupt2010sequential, kar2013cal}. Additionally, tasks in virtual environments, such as those running on simulators, often disregard observation costs because the optimization goals—such as maximizing rewards or achieving specific objectives—are not inherently aligned with the expenses involved in acquiring observations~\cite{paudel2022learning, haider2021domain}. This disconnect allows for the assumption of complete knowledge of the environment, but ignoring observation costs in such models makes reinforcement learning applications diverge from real-world practice.

\begin{figure*}[t!]
    \centering
    \includegraphics[width=\linewidth]{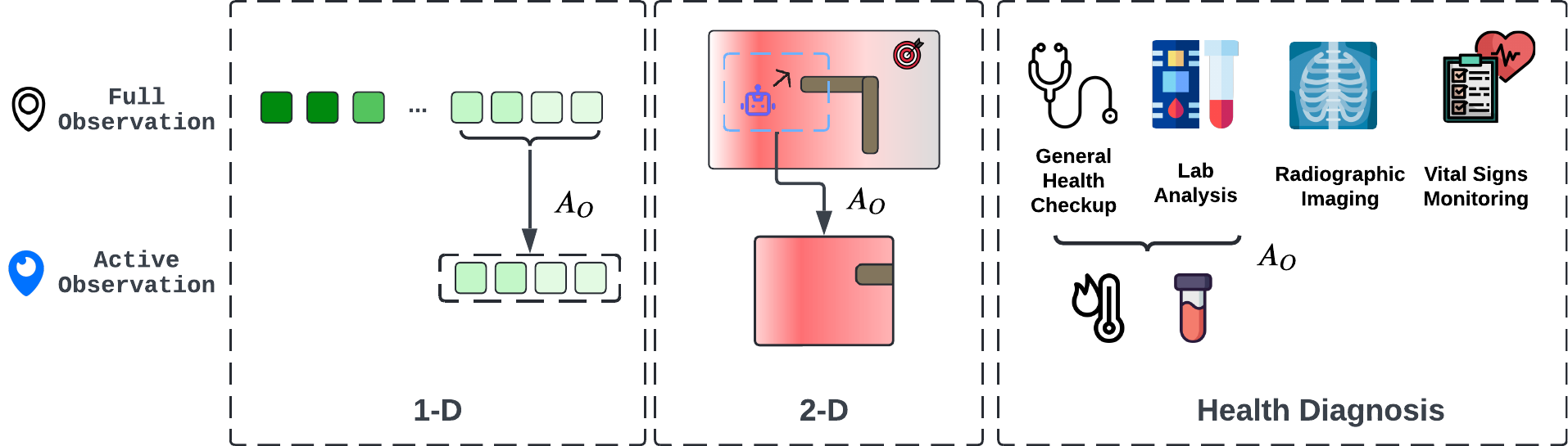}
    \caption{Active observations within full state space}
    \label{fig:state}
    \vspace{-10pt}
\end{figure*}

In many real-world applications, particularly in healthcare, decision-making processes must account for the costs associated with actively obtaining observations. Medical assessments, diagnostic tests, and patient monitoring not only require financial resources but also demand significant time from healthcare professionals and patients alike. This inherent cost associated with information gathering necessitates strategies that judiciously balance the need for information with the resources available~\cite{chen2015deep, yu2009active}. Medical assessments can be regarded as sequential decision-making problems, where treatments are administered based on the patient's current health states. These health states are inferred from observations, including physical examinations and clinical metrics. The challenge lies in making optimal decisions with limited and costly observations, which is critical for both patient outcomes and resource management~\cite{killian2020empirical, riachi2021challenges, datta2021reinforcement}. Some examples of active observation actions within the full space are presented in Figure~\ref{fig:state}.

This paper addresses the challenge of simultaneously learning an observation strategy and a control strategy in environments where observations are costly. In such settings, there is a fundamental trade-off between the cost of acquiring observations and the benefit they provide in making informed control decisions. We consider the simplest case of observation action design, where each observation action is a binary decision: whether to acquire a specific observation or not. This formulation leads to a total action space of $2^{|\mathcal{O}|} \times |\mathcal{A}_{\text{control}}|$, where $|\mathcal{O}|$ is the number of possible observations to make and $|\mathcal{A}_{\text{control}}|$ is the number of control actions. This exponential growth in the action space introduces the \textit{curse of dimensionality}, making the learning process significantly more complex than when learning a control policy alone.

To tackle this challenge, we propose an iterative, model-free deep reinforcement learning approach that decomposes the sensing and control policies. By separating these two aspects, the learning algorithm can more efficiently navigate the enlarged action space. The model focuses on learning when and what to observe, alongside determining the optimal control actions, without requiring a complete model of the environment's dynamics~\cite{silver2010monte, kaelbling1998planning}. 

We demonstrate the effectiveness of our method on two fronts of medical practice: a simulated \textit{Diagnostic Chain} task designed to capture the essential elements of the problem, and a realistic healthcare environment using \textit{HeartPole}~\cite{liventsev2021towards}. These experiments showcase the model's ability to make cost-effective observation decisions while achieving desirable control outcomes.




\section{Problem Formalism}

We address environments where the agent must actively decide not only on control actions but also on whether to acquire observations, each potentially incurring a cost. This scenario extends the classic Partially Observable Markov Decision Process (POMDP) by allowing the agent's policy to influence the observability of the environment's state. We formalize this setting as an \textit{Observation Constrained Markov Decision Process (OCMDP)}. 

An OCMDP is defined by the tuple $\mathcal{M} = (\mathcal{S}, \mathcal{A}, \mathcal{O}, \mathcal{T}, \mathcal{Z}, \mathcal{R}, \mathcal{C}, \gamma)$, where: $\mathcal{S}$ is the full state space, which can be any measurable set (not necessarily a subset of $\mathbb{R}^n$).
$\mathcal{A} = \mathcal{A}^c \times \mathcal{A}^o$ is the composite action space, consisting of control actions $\mathcal{A}^c$ and observation actions $\mathcal{A}^o$.
$\mathcal{O}$ is the set of possible observations, augmented with a null observation $\emptyset$ indicating no observation.
$\mathcal{T}: \mathcal{S} \times \mathcal{A}^c \rightarrow \mathcal{M}(\mathcal{S})$ is the state transition function, where $\mathcal{M}(\mathcal{S})$ denotes the set of probability distributions over $\mathcal{S}$.
$\mathcal{Z}: \mathcal{S} \times \mathcal{A}^o \rightarrow \mathcal{M}(\mathcal{O})$ is the observation function, dependent on both the state and the observation action.
$\mathcal{R}: \mathcal{S} \times \mathcal{A}^c \rightarrow \mathbb{R}$ is the reward function.
$\mathcal{C}: \mathcal{A}^o \rightarrow \mathbb{R}_{\geq 0}$ is the cost function associated with making observations.
$\gamma \in [0,1)$ is the discount factor.

At each time step $t$, the agent proceeds as follows: First, the agent selects an observation action $a_t^o \in \mathcal{A}^o$ using the observation policy $\pi^o$, based on its current history $h_t$ or belief state $b_t$: $a_t^o \sim \pi^o(a_t^o \mid h_t).$ Then, the agent receives an observation $o_t$ based on the observation function, which depends on the current full state $s_t$ (might be unobservable) and the selected observation action $a_t^o$: $o_t \sim \mathcal{Z}(o_t \mid s_t, a_t^o).$
The agent updates its history $h_t$ to include the new observation $o_t$, potentially updating its belief state $b_t$. Next, the agent selects a control action $a_t^c \in \mathcal{A}^c$ using the control policy $\pi^c$, conditioned on the updated history or belief state: $a_t^c \sim \pi^c(a_t^c \mid h_t, o_t).$ The environment transitions from state $s_t$ to $s_{t+1}$ according to the state transition function: $s_{t+1} \sim \mathcal{T}(s_{t+1} \mid s_t, a_t^c).$ Simultaneously, the agent obtains a reward $r_t = \mathcal{R}(s_t, a_t^c)$ and incurs an observation cost $c_t = \mathcal{C}(a_t^o)$. The total immediate utility at time $t$ is then $u_t = r_t - c_t$.

The agent's objective is to find a policy $\pi = (\pi^o, \pi^c)$ that maximizes the expected cumulative discounted utility:
\begin{equation}
\label{eqn:learning_obj}
\max_{\pi} J(\pi) = \max_{\pi^o, \pi^c} \mathbb{E}_{\tau\sim\pi} \left[ \sum_{t=0}^{\infty} \gamma^t (r_t - c_t) \right],
\end{equation}
where the expectation is over trajectories induced by the policy $\pi$, and the policies map histories to action distributions:
\[
\pi^o(a_t^o \mid h_t), \quad \pi^c(a_t^c \mid h_t, o_t),
\]
with history $h_t = \{ o_0, a_0^o, a_0^c, r_0, c_0, \ldots, o_{t-1}, a_{t-1}^o, a_{t-1}^c, r_{t-1}, c_{t-1} \}.$

Since the full state at time $t$, i.e., $s_t$, is costly to observe, the agent may maintain a belief state $b_t \in \mathcal{M}(\mathcal{S})$, which is updated based on the history $h_t$ and encapsulates the agent's knowledge about the environment.

\section{Related Work}

\paragraph{POMDP}
Generally speaking, Partially Observed Markov Decision Process (POMDP,~\cite{spaan2012partially,krishnamurthy2016partially}) is the special case of the setting considered in this work, as the latent system state (e.g., the state of the patient) can only be inferred based on partial observations, However, distinguished from vanilla POMDP problems where the observation space is fixed, we consider the observation as an \textit{active decision} that can varies across states. From this perspective, our problem setting reduces to POMDP when a trivial \textit{always observe} strategy is applied. 


\paragraph{Active Sensing}

Active sensing is a framework where an agent actively decides which observations to acquire to improve task performance while minimizing the associated costs~\cite{yoon2019asac}. In environments where observations are costly or resource-intensive, active sensing enables agents to selectively focus on the most informative data. Our work relates to active sensing in that setting the discount factor $\gamma$ to zero and defining the reward function $\mathcal{R}$ as a classifier for a downstream task reduces our learning objective in Eqn.~(\ref{eqn:learning_obj}) to the traditional active sensing task. However, our OCMDP solver extends beyond conventional active sensing by incorporating sequential decision-making over time. Unlike standard active sensing, which typically involves single-step decisions in static environments, our approach addresses the challenges of making both action and observation decisions in dynamic environments across multiple time steps. By considering a positive discount factor $\gamma > 0$, we balance immediate observation costs with long-term benefits, enabling a strategy that accounts for the future impact of current decisions. This integration allows us to tackle more complex scenarios where traditional active sensing methods may not be feasible.



\paragraph{Model-Based Methods}

Model-based reinforcement learning (RL) approaches incorporate a model of the environment's dynamics into the policy learning process, enabling the agent to plan and make informed decisions based on predicted future states~\cite{sutton2018reinforcement}. Recently, several studies have integrated sensing or observation strategies into the policy learning framework. Additionally, \citet{yin2020reinforcement} introduces the task of Active Feature Acquisition, which emphasizes representation learning. Their approach utilizes Variational Autoencoders (VAEs) for imputing missing features and inferring latent states, operating within the broader category of model-based RL. These model-based methods typically require accurate models of the environment or observation processes, which can be difficult to obtain in complex or high-dimensional settings. In contrast, our approach adopts a model-free reinforcement learning methodology, circumventing the need for explicit environment models. By leveraging recent advances in model-free deep RL, our method can be more scalable and adaptable to diverse and intricate environments where modeling the dynamics is impractical or infeasible. This allows our OCMDP solver to effectively handle the uncertainties and complexities inherent in partially observable settings without relying on predefined models.

\paragraph{ML-enhanced Decision Making in Healthcare}

Machine learning (ML) has significantly advanced decision-making processes in healthcare by enabling more accurate diagnoses, personalized treatment plans, and efficient resource allocation~\cite{esteva2019guide, rajkomar2019machine}. Supervised learning techniques, such as deep neural networks, have been employed to analyze medical imaging data, leading to breakthroughs in detecting conditions like diabetic retinopathy and skin cancer with performance comparable to or exceeding that of human experts~\cite{gulshan2016development, esteva2017dermatologist}. Additionally, reinforcement learning (RL) has been applied to optimize treatment strategies for chronic diseases, where the complexity and variability of patient responses make traditional decision-making approaches inadequate~\cite{komorowski2018opportunities, nair2020reinforcement}. To further enhance the considerations over long-term planning, our proposed approach, OCMDP, adopts deep reinforcement learning (DRL), which addresses the complexities of balancing observation costs with decision-making benefits in more general healthcare cases. By simultaneously learning observation and control strategies, OCMDP enhances the efficiency and personalization of healthcare delivery, enabling more adaptive and cost-effective resource management. These advancements underscore the transformative potential of ML in enhancing the efficiency, effectiveness, and personalization of healthcare delivery.

\section{The Model-Free Approach}

We propose a \textbf{model-free framework} to address the Active Observation Markov Decision Process (AOMDP) problem. This framework involves two distinct policies: an \textbf{observation policy} $\pi^o$ and a \textbf{control policy} $\pi^c$, which together form the combined policy $\pi = (\pi^o, \pi^c)$. We also plot out the diagram of our Observation-Constrained MDP (OCMDP) solver in Figure ~\ref{fig:ocmdp}.

\subsection{Trajectory-Based Action-Value Function}

We define the \textbf{trajectory-based action-value function} $Q^\pi(h_t, a_t)$ as:

\begin{equation}
\label{eq:trajectory_action_value}
Q^\pi(h_t, a_t) = \mathbb{E}_{\tau \sim \pi} \left[ \sum_{k=0}^\infty \gamma^k \left( r(s_{t+k}, a^c_{t+k}) - \mathbf{c}^\top a^o_{t+k} \right) \right],
\end{equation}

$h_t$ is the history up to time $t$, $a_t$ is the action taken at time $t$, $\gamma \in [0,1)$ is the discount factor, $r(s, a^c)$ represents the reward for taking control action $a^c$ in state $s$, $\mathbf{c}$ is the cost vector associated with observation actions, $a^o$ denotes the observation actions. This function captures the expected cumulative reward minus the cumulative observation cost over an infinite horizon. For analytical convenience, we decompose $Q^\pi(h_t, a_t)$ into two components:

\begin{align*}
Q^\pi(h_t, a_t) &= R^\pi(h_t, a_t) - C^\pi(h_t, a_t), \\
\text{$where$} \quad
R^\pi(h_t, a_t) &= \mathbb{E}_{\pi} \left[ \sum_{k=0}^\infty \gamma^k r(s_{t+k}, a^c_{t+k}) \right], \\
C^\pi(h_t, a_t) &= \mathbb{E}_{\pi} \left[ \sum_{k=0}^\infty \gamma^k \mathbf{c}^\top a^o_{t+k} \right].
\end{align*}

Here, $R^\pi(h_t, a_t)$ represents the expected cumulative reward, and $C^\pi(h_t, a_t)$ represents the expected cumulative observation cost.

\subsection{Iterative Optimization}
\label{sec:iter}

Our approach employs a model-free, iterative optimization framework that alternates between refining the control and observation policies to achieve optimal performance. This methodology leverages the interplay between control actions and observation strategies, ensuring that each policy enhancement step contributes to the overall improvement of the system's behavior.

Initially, we focus on optimizing the control policy while keeping the observation policy fixed. By holding the observation policy $\overline{\pi}^o$ constant, we isolate the control dynamics and aim to maximize the expected cumulative reward. The control-specific action-value function is defined as:

\begin{equation}
\label{eq:control_value_function}
Q^{\pi^c}_{\overline{\pi}^o}(h_t, a_t) = \mathbb{E}_{\pi^c, \overline{\pi}^o} \left[ \sum_{k=0}^\infty \gamma^k r_o(s_{t+k}, a^c_{t+k}) \right],
\end{equation}

where the modified reward incorporates observation costs:$r_o(s, a^c) = r(s, a^c) - \mathbf{c}^\top \overline{a}^o.$

Here, $\overline{a}^o$ represents the observation actions dictated by the fixed observation policy $\overline{\pi}^o$. By optimizing the control policy $\pi^c$, we ensure that the agent selects actions that not only maximize rewards but also account for the costs associated with observations, effectively embedding the observation strategy within the control dynamics.

Subsequently, we shift our focus to optimizing the observation policy while keeping the control policy $\overline{\pi}^c$ fixed. The objective here is to minimize the expected cumulative observation costs without compromising the reward structure established by the control policy. The observation-specific action-value function is articulated as:

\begin{equation}
\label{eq:observation_value_function}
Q^{\pi^o}_{\overline{\pi}^c}(h_t, a_t) = \mathbb{E}_{\pi^o, \overline{\pi}^c} \left[ \sum_{k=0}^\infty \gamma^k r_c(s_{t+k}, a^o_{t+k}) \right],
\end{equation}

where the reward is adjusted based on the control actions: $r_c(s, a^o) = r(s, \overline{a}^c) - \mathbf{c}^\top a^o.$

In this context, $\overline{a}^c$ denotes the control actions determined by the fixed control policy $\overline{\pi}^c$. Optimizing the observation policy $\pi^o$ under these conditions ensures that the agent selectively acquires information, balancing the trade-off between the cost of observations and the benefits they provide in enhancing decision-making.

This iterative process of alternating between control and observation policy optimization is motivated by the need to decouple the complexities inherent in each policy type. By isolating the optimization steps, we can more effectively navigate the high-dimensional policy spaces, ensuring that improvements in one policy component do not inadvertently degrade the performance of the other. Moreover, this separation facilitates targeted enhancements, allowing for more nuanced adjustments that align with the specific objectives of control and observation strategies.

\subsection{Policy Gradient Optimization}
\begin{figure*}[t!]
    \centering
    \includegraphics[width=0.8\linewidth]{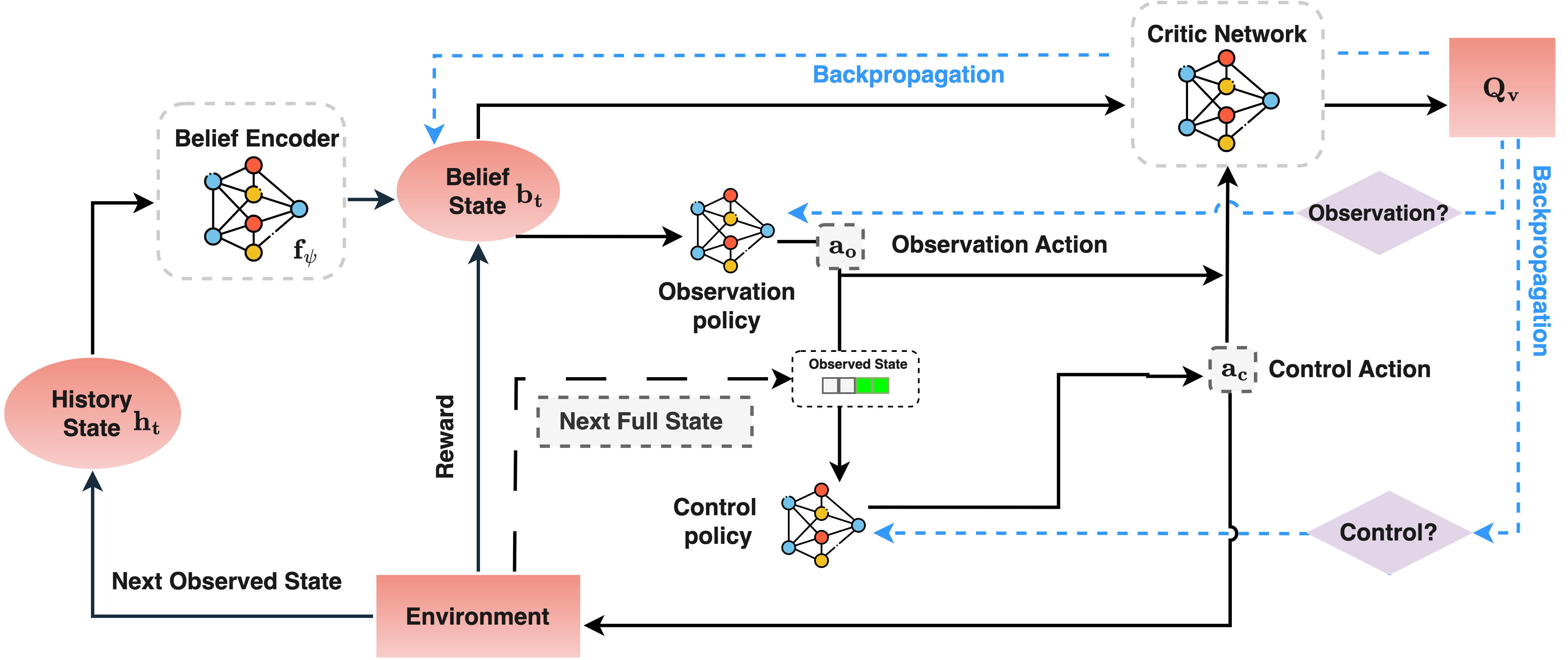}
    \caption{Diagram of Observation-Constrained MDP (OCMDP) Solver}
    \label{fig:ocmdp}
\end{figure*}
To operationalize the iterative policy optimization framework, we adopt policy gradient methods, which are well-suited for optimizing parameterized policies in continuous and high-dimensional spaces. We parameterize the control policy with parameters $\theta$ and the observation policy with parameters $\phi$, enabling flexible and scalable policy representations.

The optimization objectives for both policies are defined to align with their respective goals of maximizing rewards and minimizing observation costs. For the control policy, the objective is to maximize the expected cumulative reward, which is formalized as:

\begin{equation}
\label{eq:control_objective}
\mathcal{J}(\theta) = \mathbb{E}_{\pi^c_\theta, \overline{\pi}^o} \left[ Q^{\pi^c_\theta}_{\overline{\pi}^o}(h_t, a_t) \right].
\end{equation}

The gradient of this objective with respect to $\theta$ is derived using the policy gradient theorem:

\begin{equation}
\nabla_\theta \mathcal{J}(\theta) = \mathbb{E}_{\pi^c_\theta, \overline{\pi}^o} \left[ \nabla_\theta \log \pi^c_\theta(a_t^c \mid h_t) \cdot Q^{\pi^c_\theta}_{\overline{\pi}^o}(h_t, a_t) \right].
\end{equation}

This gradient expression guides the adjustment of $\theta$ in a direction that increases the expected cumulative reward, effectively enhancing the control policy's performance.

Conversely, the observation policy aims to minimize the expected cumulative observation costs while maintaining the reward structure influenced by the control policy. This objective is encapsulated as:

\begin{equation}
\label{eq:observation_objective}
\mathcal{J}(\phi) = -\mathbb{E}_{\pi^o_\phi, \overline{\pi}^c} \left[ Q^{\pi^o_\phi}_{\overline{\pi}^c}(h_t, a_t) \right].
\end{equation}

The negative sign indicates the minimization of observation costs. The corresponding gradient with respect to $\phi$ is given by:

\begin{equation}
\nabla_\phi \mathcal{J}(\phi) = -\mathbb{E}_{\pi^o_\phi, \overline{\pi}^c} \left[ \nabla_\phi \log \pi^o_\phi(a_t^o \mid h_t) \cdot Q^{\pi^o_\phi}_{\overline{\pi}^c}(h_t, a_t) \right].
\end{equation}

This gradient facilitates the reduction of observation costs by adjusting $\phi$ in a direction that decreases the expected cumulative costs, thereby optimizing the observation policy's efficiency.

The integration of these policy gradient updates within the iterative optimization framework ensures that both control and observation policies are incrementally improved in a coordinated manner. This dual optimization strategy not only enhances the agent's ability to make informed and cost-effective decisions but also fosters a balanced trade-off between reward maximization and cost minimization. The resulting policies are thus finely tuned to navigate the complexities of the environment, leveraging observations judiciously to achieve superior performance.


\subsection{Theoretical Analysis of Iterative Policy Optimization}

As discussed in section~\ref{sec:iter}, our COMDP solver employs a sequential optimization strategy, grounded in dynamic programming principles, to iteratively refine both control and observation strategies. The following sections provide the theoretical foundation supporting this approach.

\begin{relemma}[Value Function Contraction] 
\label{lemma:1} 
Consider the Bellman optimality operator $\mathcal{T}$ defined as:
\[
\mathcal{T}Q(h_t, a_t) = r(s_t, a_t) + \gamma \mathbb{E}_{s_{t+1}} \left[ \max_{a_{t+1}} Q(h_{t+1}, a_{t+1}) \right].
\]
Under the supremum norm $\|Q\|_\infty = \sup_{h_t, a_t} |Q(h_t, a_t)|$, the operator $\mathcal{T}$ acts as a \textbf{contraction mapping} with a contraction factor of $\gamma$. Repeated application of $\mathcal{T}$ guarantees convergence to the unique optimal action-value function $Q^*$.    
\end{relemma}

\begin{proof}
To establish that $\mathcal{T}$ is a contraction, consider any two value functions $Q_1$ and $Q_2$. We aim to show that:
\[
\|\mathcal{T}Q_1 - \mathcal{T}Q_2\|_\infty \leq \gamma \|Q_1 - Q_2\|_\infty.
\]
Starting from the definition of the supremum norm:
\begin{align*}
\|\mathcal{T}Q_1 - \mathcal{T}Q_2\|_\infty &= \sup_{h_t, a_t} \left| \mathcal{T}Q_1(h_t, a_t) - \mathcal{T}Q_2(h_t, a_t) \right| \\
&= \gamma \sup_{h_t, a_t} \left| \mathbb{E}_{s_{t+1}} \left[ \max_{a_{t+1}} Q_1(h_{t+1}, a_{t+1}) - \max_{a_{t+1}} Q_2(h_{t+1}, a_{t+1}) \right] \right| \\
&\leq \gamma \sup_{h_t, a_t} \mathbb{E}_{s_{t+1}} \left| \max_{a_{t+1}} Q_1(h_{t+1}, a_{t+1}) - \max_{a_{t+1}} Q_2(h_{t+1}, a_{t+1}) \right| \\
&\leq \gamma \|Q_1 - Q_2\|_\infty.
\end{align*}
The first inequality follows from the linearity of expectation and the properties of the supremum norm. The second inequality leverages the fact that the maximum difference between $Q_1$ and $Q_2$ across all actions bounds the expectation. This demonstrates that $\mathcal{T}$ reduces the distance between any two value functions by at least a factor of $\gamma$. By Banach's Fixed Point Theorem, this contraction property ensures that repeated application of $\mathcal{T}$ will converge to the unique fixed point $Q^*$, the optimal action-value function.
\end{proof}

\begin{reremark}
A similar contraction property is satisfied by the \textbf{Bellman expectation operator} $\mathcal{T}^\pi$:
\[
\mathcal{T}^\pi Q(h_t, a_t) = r(s_t, a_t) + \gamma \mathbb{E}_{s_{t+1}, a_{t+1} \sim \pi} \left[ Q(h_{t+1}, a_{t+1}) \right].
\]
This operator is fundamental in policy evaluation, ensuring that iterative application under a fixed policy $\pi$ will converge to the corresponding value function $Q^\pi$.
\end{reremark}

\begin{relemma}[Policy Enhancement]
\label{lemma:2}
Given a policy $\pi$, define an updated policy $\pi'$ such that for every history $h_t$:
\[
\pi'(h_t) = \arg\max_{a_t} Q^\pi(h_t, a_t).
\]
Then, the value function under $\pi'$ satisfies:
\[
Q^{\pi'}(h_t, a_t) \geq Q^\pi(h_t, a_t), \quad \forall h_t, a_t.
\]
\end{relemma}

\begin{proof}
By the definition of the updated policy $\pi'$, for each history $h_t$, the action $\pi'(h_t)$ is chosen to maximize the expected return as per the current value function $Q^\pi$. Formally:
\[
Q^\pi(h_t, \pi'(h_t)) = \max_{a_t} Q^\pi(h_t, a_t).
\]
This implies that:
\[
Q^\pi(h_t, \pi'(h_t)) \geq Q^\pi(h_t, \pi(h_t)),
\]
since $\pi$ is just one of the possible policies and $\pi'$ selects the action with the highest $Q$-value. Consequently, when we evaluate the new policy $\pi'$, its value function $Q^{\pi'}$ must satisfy:
\[
Q^{\pi'}(h_t, a_t) \geq Q^\pi(h_t, a_t).
\]
This inequality holds for all histories $h_t$ and actions $a_t$, demonstrating that the policy enhancement step results in a policy that is at least as good as the original policy $\pi$ in terms of expected return.
\end{proof}

\begin{reproposition}[Convergence to a Locally Optimal Policy]
\label{remark:3}
By alternately applying the value function contraction and policy enhancement steps described in Lemmas~\ref{lemma:1} and \ref{lemma:2}, the iterative process converges to a locally optimal joint policy $\overline{\pi}^* = (\overline{\pi}^{c,*}, \overline{\pi}^{o,*})$. This locally optimal policy satisfies the condition:
\[
Q^{\overline{\pi}^*}(h_t, a_t) \geq Q^{\pi_i}(h_t, a_t) \geq \dots \geq Q^{\pi_1}(h_t, a_t) \geq Q^{\pi_0}(h_t, a_t), \quad \forall h_t, a_t,
\]

where $\pi_i$ represents the policy obtained after the $i$-th iteration. This sequence of policies forms a monotonically improving trajectory in the space of policies, ensuring that each successive policy is at least as good as its predecessor in terms of the expected return.
\end{reproposition}

\begin{reproposition}[Conditions for Global Optimality]
\label{remark:4}
The globally optimal joint policy $\pi^* = (\pi^{c,*}, \pi^{o,*})$ can be attained provided that either the control policy component $\overline{\pi}^{c,*}$ converges to its optimal counterpart $\pi^{c,*}$ or the observation policy component $\overline{\pi}^{o,*}$ converges to $\pi^{o,*}$. 

\end{reproposition}

\begin{proof}
Assume that the control policy component $\overline{\pi}^{c,*}$ converges to the optimal control policy $\pi^{c,*}$. Under this condition, optimizing the observation policy $\pi^o$ will naturally lead it to converge towards the optimal observation policy $\pi^{o,*}$. Conversely, if the observation policy does not converge to $\pi^{o,*}$, there exists an alternative observation policy $\pi^{o} \neq \pi^{o,*}$ such that:
\[
Q^{\pi^{o}}(h_t, a_t) > Q^{\pi^{o,*}}(h_t, a_t),
\]
which contradicts the assumption that $\pi^{o,*}$ is the optimal observation policy as defined by the objective in Eqn.~(\ref{eq:observation_objective}). A symmetric argument holds if we assume that the observation policy component $\overline{\pi}^{o,*}$ converges to $\pi^{o,*}$. Thus, ensuring the optimality of either the control or the observation policy suffices to drive the joint policy towards global optimality.
\end{proof}

Although exhaustively searching the entire policy space to identify either $\pi^{c,*}$ or $\pi^{o,*}$ is computationally infeasible, Proposition~\ref{remark:4} provides a crucial insight: achieving the optimality of either the control policy or the observation policy significantly contributes to the convergence of the overall policy towards a global optimum. This observation underscores the importance of strategically prioritizing the optimization of one policy component to facilitate the convergence of the other.

Given the inherent difficulty in optimizing the observation policy without a robust control policy, our approach strategically prioritizes establishing a high-performing control policy initially. This prioritization is motivated by the fact that a well-optimized control policy can effectively guide the acquisition of relevant observations, thereby reducing the complexity of optimizing the observation policy subsequently.

\subsection{Implementing the Iterative Approach}


To operationalize our sequential optimization strategy, we introduce a learning framework that decouples the optimization of the control policy from the representation learning, specifically the belief state inference. This framework leverages a shared belief state module to facilitate efficient and scalable learning. The architecture comprises three neural networks parameterized by $\psi$, $\phi$, and $\theta$, corresponding to belief state extraction, observation policy, and control policy, respectively, details of the algorithmic implementations are presented in Algorithm~\ref{alg:aomdp}.

\begin{algorithm}[htbp]
\SetAlgoLined
\caption{Iterative Policy Optimization for OCMDPs}
\label{alg:aomdp}
\kwInit{\\
    Initialize belief state extractor parameters $\psi$;\\
    Initialize observation policy parameters $\phi$ (curiosity-driven);\\
    Initialize control policy parameters $\theta$;
}
\Repeat{Convergence}{
    \tcc{\textbf{Derive Belief State and Determine Actions}}
    Derive belief state: $b_t = f_{\psi}(h_t)$\;
    Determine observation action: $a^o_t \sim \pi^o_{\phi}(\cdot \mid b_t)$\;
    Obtain observation based on $a^o_t$: $o_t = \mathcal{O}(s_t, a^o_t)$\;
    Determine control action: $a^c_t \sim \pi^c_{\theta}(\cdot \mid o_t)$\;
    
    \tcc{\textbf{Control Policy Optimization}}
    Fix observation policy $\pi^o_{\phi}$\;
    Optimize $\psi$ and $\theta$ to maximize expected cumulative reward:\;
    \Indp $\psi, \theta \leftarrow \arg\max_{\psi, \theta} \; \mathbb{E}_{\pi^c_{\theta}, \pi^o_{\phi}} \left[ \sum_{t=0}^{\infty} \gamma^t \; r(s_t, a^c_t) \right]$\;
    \Indm
    
    \tcc{\textbf{Observation Policy Optimization}}
    Fix control policy $\pi^c_{\theta}$\;
    Optimize $\psi$ and $\phi$ to minimize observation costs and belief state error:\;
    \Indp $\psi, \phi \leftarrow \arg\min_{\psi, \phi} \; \mathbb{E}_{\pi^o_{\phi}, \pi^c_{\theta}} \left[ \sum_{t=0}^{\infty} \gamma^t \left( \mathbf{c}^\top a^o_t + \lambda \; \mathcal{L}_{\text{belief}}(b_t, s_t) \right) \right]$\;
    \Indm
    
    \tcc{\textbf{Optional Fine-Tuning of Control Policy}}
    Optimize $\theta$ to adapt to the updated observation policy:\;
    \Indp $\theta \leftarrow \arg\max_{\theta} \; \mathbb{E}_{\pi^c_{\theta}, \pi^o_{\phi}} \left[ \sum_{t=0}^{\infty} \gamma^t \; r(s_t, a^c_t) \right]$\;
    \Indm
}
\KwRet{Optimized policies $\pi^c_{\theta}$ and $\pi^o_{\phi}$;}
\end{algorithm}

At time $t$, the belief state $b_t$ is derived from the history $h_t$ through the function $b_t = f_\psi(h_t)$. The observation action $a^o_t$ is then determined by the observation policy $\pi^o_\phi$, parameterized by $\phi$, and conditioned on the belief state: $a^o_t \sim \pi^o_\phi(\cdot \mid b_t)$. The observation action $a^o_t$ specifies which components of the environment to observe, resulting in an observation $o_t$, which is a function of the true state $s_t$ and the observation action $a^o_t$: $o_t = \mathcal{O}(s_t, a^o_t)$, where $\mathcal{O}$ represents the observation function that returns the observed components of the state based on $a^o_t$. The control action $a^c_t$ is then determined by the control policy $\pi^c_\theta$, parameterized by $\theta$, and conditioned on the observed state $o_t$: $a^c_t \sim \pi^c_\theta(\cdot \mid o_t)$. This process ensures that the control actions are based on the most recent observations, which have been acquired according to the observation policy. The observation policy aims to acquire information that is most beneficial for decision-making while minimizing the associated costs.

The optimization process proceeds iteratively, following a structured sequence of updates. Initially, the observation policy is initialized using a curiosity-driven approach, which encourages the gathering of comprehensive information without immediate consideration of costs. This initialization ensures that the belief state extractor has access to rich information, crucial for accurately inferring the state. Subsequently, we optimize the parameters $\psi$ and $\theta$ to maximize the expected return using the currently accessible information, resulting in the optimized parameters $\psi_1^*$ and $\theta_1^*$, and the most accurate accessible state representation $b^* = f_{\psi_1^*}(h_t)$. The control policy $\pi^c_{\theta_1^*}$ is thus optimized based on the observations $o_t$ resulting from the observations acquired via the observation policy.

In the next iteration, while keeping the control policy fixed, we optimize $\psi$ and $\phi$ to minimize state inference costs. Since the control policy remains constant, these optimized components should recover the optimal belief state $b^*$ as accurately as possible while minimizing observation costs. This involves updating the observation policy $\pi^o_\phi$ to select observation actions that are most informative for the control policy while incurring minimal cost. Once this is achieved, we can fine-tune the control policy. This iterative process continues until convergence, with each iteration refining the observation and control policies to balance the trade-off between observation costs and control performance. The implementation details and reinforcement learning loops are shown in Figure ~\ref{fig:ocmdp}.

\section{Experiments}

\subsection{The Diagnostic Chain}
\paragraph{Design of the Environment}

\begin{figure*}[htbp]
    \centering
    \includegraphics[width=0.8\linewidth]{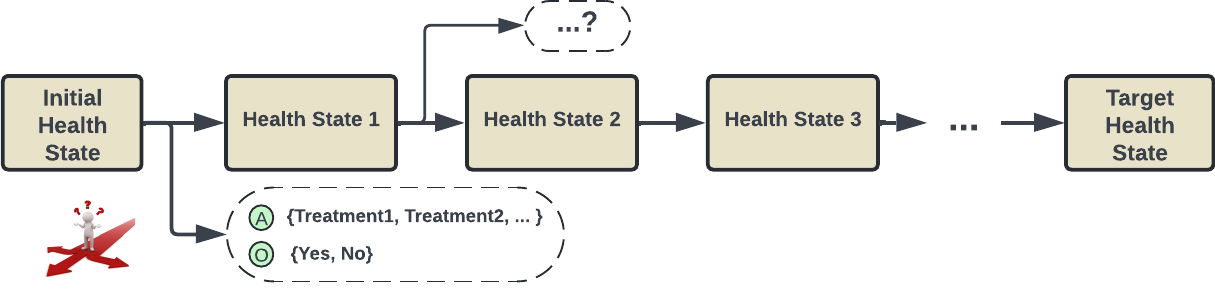}
    \caption{Diagnostic Task}
    \label{fig:toy_task}
\end{figure*}

We validate our approach with a simple and carefully-designed \textit{Diagnostic Chain} task in which an agent is required to transition a patient to a \textbf{target health state} within a sequential chain of health states. Each node in this chain represents a distinct health condition of a patient. For the purposes of this experiment, we assume that all patients consistently begin at the \textbf{baseline health state}, depicted as the leftmost node in the chain. The agent has access to multiple treatment options, which can be interpreted as different medical prescriptions. Specifically, we consider two possible treatment actions, denoted as $\mathcal{A}^T = \{\text{Treatment}_1, \text{Treatment}_2\}$, representing the control actions available to the agent. Additionally, the agent can choose between two observation actions, $\mathcal{A}^O = \{\text{Observe}, \text{Not Observe}\}$, which determine whether the agent will assess the effectiveness of the administered treatment. The MDP chain of such proposed task is presented in Figure~\ref{fig:toy_task}.

Each treatment action incurs a cost, represented by $-C_T$, where $C_T > 0$, while choosing to perform an observation incurs a cost of $-C_O$, where $C_O > 0$. Successfully reaching the target health state yields a positive reward, denoted by $R_T$, where $R_T > 0$. The effectiveness of each treatment varies across different patients. For each episode, one of the two treatment actions will successfully transition the patient to the next health state in the chain with an equal probability of $50\%$. Conversely, there is a $50\%$ probability that the treatment will not alter the patient's current health state.

The optimal strategy for the agent involves selectively performing observations to determine the effectiveness of a treatment before deciding whether to continue with the same treatment or switch to an alternative. Initially, the agent selects a treatment action without performing an observation. After administering the first treatment, the agent may choose to observe the outcome. If the observation action is selected (\textit{Observe}), the agent incurs the associated cost $-C_O$ and gains information about the treatment's effectiveness. An effective treatment leads the agent to continue with the same treatment, whereas an ineffective treatment prompts the agent to switch to the alternative treatment. This approach mirrors real-world healthcare decision-making, where clinicians may perform diagnostic tests to evaluate treatment efficacy before making subsequent treatment decisions.

If the agent chooses not to perform an observation action (\textit{Not Observe}), it forgoes the additional cost and does not receive explicit information about the treatment's effectiveness. In this scenario, the agent must rely on \textbf{blurred state information}, meaning that its perception of the current health state is noisy or incomplete. This blurred state ensures consistency in the input state representations, even in the absence of direct observations, thereby testing the agent's ability to make informed decisions under uncertainty.

The agent continues to alternate between treatment and observation actions, adhering to its policy until it successfully reaches the target health state. Upon reaching the target, the agent receives the reward $R_T$, concluding the episode. This setup challenges the agent to balance the costs associated with treatments and observations against the benefits of achieving optimal patient health outcomes. By incorporating uncertainty in treatment effectiveness and the option to perform costly observations, the framework encourages the development of efficient and effective strategies for sequential decision-making under uncertainty.

\paragraph{Instantiation and control action}
In our experiments, we use Markov chains of length $N=5$ and set the terminal reward $R_T=25$ to ensure positive cumulative returns for optimal policies. Each dimensional observation $i$ incurs a cost of $C_{O_i} = -0.4$, while each control action incurs a cost of $C_T = -1$. These settings align with the average therapy situation from the real-world Medkit-Learn medical simulator~\cite{chan2021medkitlearn}. We expand the action space from the basic binary choice to $M=6$ possible actions, where only one action correctly progresses toward the target state on the chain. Each episode is capped at a maximum length of 8 steps to ensure timely decision-making.

\paragraph{Baseline Methods}

Our experiments compare our proposed approach against several widely-used model-free reinforcement learning (RL) algorithms, including Proximal Policy Optimization (PPO,~\cite{schulman2017proximal}), Soft Actor-Critic (SAC~\cite{haarnoja2018soft}), and Deep Deterministic Policy Gradient (DDPG~\cite{lillicrap2015continuous}). Additionally, to validate the core concepts of our OCMDP solver, we evaluate our method under different observational settings: \textbf{always observe}, \textbf{never observe}, and \textbf{optimally observe (ours)}. These varied settings enable us to assess the robustness and effectiveness of our solver in environments with differing levels of observability. This setting is also used in our second experiment, as described section~\ref{sec:heartpole}.

\begin{figure*}[t!]
    \centering
    \includegraphics[width=0.9\linewidth]{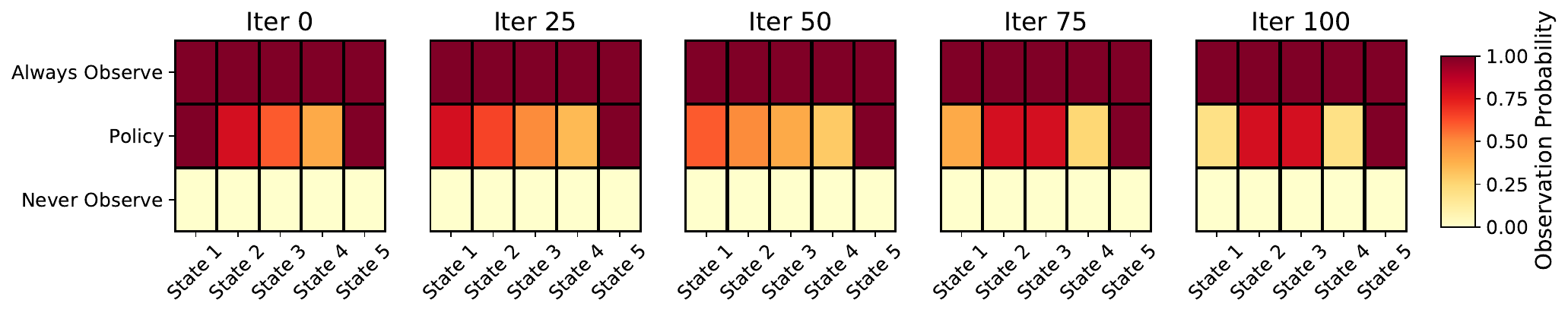}
    \caption{An Illustration of the observation policy evolution (\textit{Diagnostic Chain})}
    \label{fig:obs_evo}
\end{figure*}

\begin{figure*}[t!]
    \centering
    \includegraphics[width=0.9\linewidth]{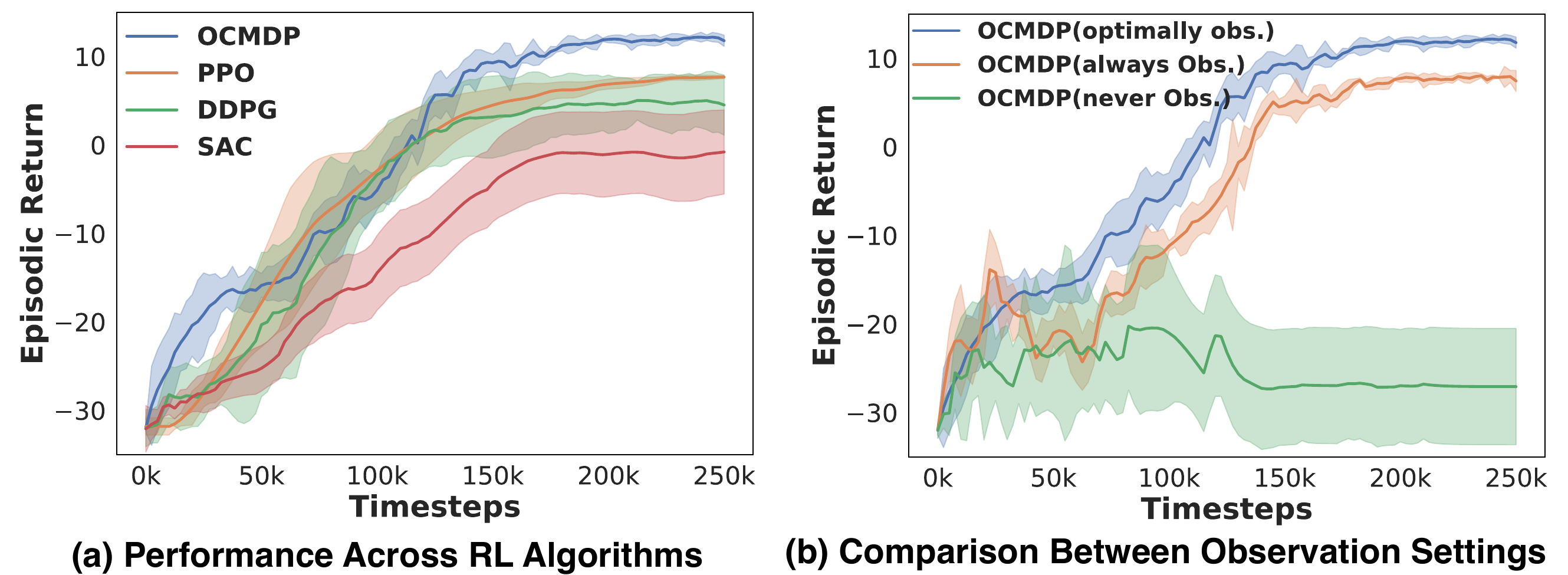}
    \caption{Performance comparison between our proposed iterative policy optimization method and baseline approaches on the OCMDPs over \textit{Diagnostic Chain} Task.}
    \label{fig:diagnostic_chain}
\end{figure*}

\subsection{Experiments on the HeartPole Healthcare Simulator}
\label{sec:heartpole}

In this study, we employ the \textit{HeartPole} environment~\cite{liventsev2021towards} for our experimental evaluations\footnote{\url{https://github.com/vadim0x60/heartpole}}. \textit{HeartPole} is a simplified, rule-based healthcare simulation developed to facilitate the benchmarking of reinforcement learning (RL) algorithms in medical contexts. Designed as an OpenAI Gym environment, \textit{HeartPole} serves as an introductory platform for testing RL techniques before applying them to more complex clinical datasets.

\textit{HeartPole} models a healthcare scenario where the agent's objective is to maximize productivity while maintaining the patient's health. The action space consists of four discrete actions: \texttt{do nothing}, \texttt{drink coffee}, \texttt{drink beer}, and \texttt{sleep}. These actions influence the patient's state, which is represented by six continuous observation variables: alertness, hypertension, intoxication, time since last sleep, time elapsed, and work done. The agent receives a reward of +1 for each unit of work completed, incentivizing productivity, and incurs a penalty of -100 if a heart attack occurs, emphasizing the importance of maintaining patient health. Each dimensional observation incurs a cost of $C_{O_i} = -0.05$. Each episode is capped at 1000 steps to balance sufficient exploration and training efficiency.

The controlled and transparent nature of \textit{HeartPole} allows researchers to systematically evaluate and refine RL algorithms within a healthcare-themed setting. By providing a clear and manageable environment, \textit{HeartPole} enables the assessment of various RL methodologies, including policy-based and value-based approaches, in achieving a balance between productivity and health maintenance. This setup is particularly advantageous for identifying strengths and limitations of different algorithms in a simplified yet relevant medical scenario before transitioning to more intricate and data-intensive environments.

Through the use of \textit{HeartPole}, we can effectively benchmark our proposed RL algorithms against established baseline methods. The discrete action space and well-defined reward structure facilitate a focused analysis of algorithm performance, stability, and convergence. Additionally, the simulation's ability to generate reproducible and interpretable results supports a thorough investigation into the decision-making processes of the agents, providing insights that are transferable to real-world clinical applications.

\subsection{Experimental Results}

\begin{figure*}[t!]
    \centering
    \includegraphics[width=0.9\linewidth]{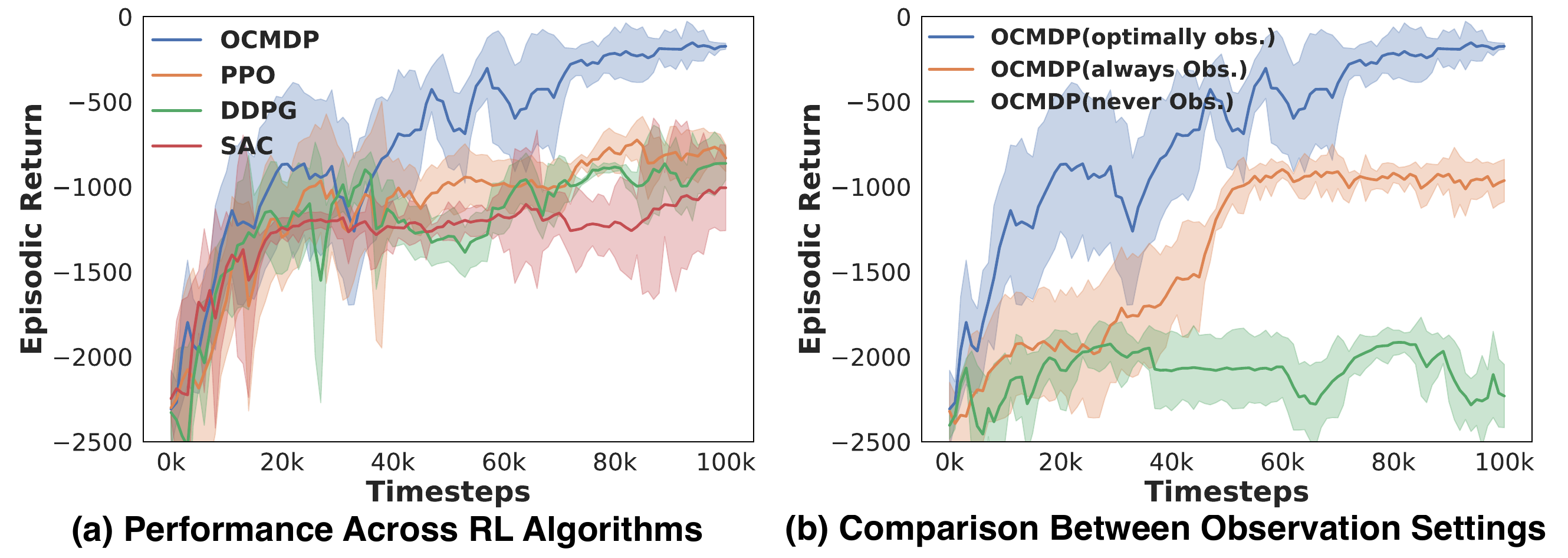}
    \caption{Performance comparison between our proposed iterative policy optimization method and baseline approaches on the OCMDPs over \textit{HeartPole} Task.}
    \label{fig:heartpole}
\end{figure*}

The experimental results are shown in Figures~\ref{fig:diagnostic_chain},~\ref{fig:heartpole} and \ref{fig:obs_evo}.  Figure~\ref{fig:obs_evo} illustrates the evolution of observation policies in our simulated \textit{Diagnostic Chain} tasks across different training steps. The visualization reveals how the policy progressively learns to balance observation costs against potential rewards, developing a strategic approach to achieve the final targets efficiently.

Figures~\ref{fig:diagnostic_chain} and~\ref{fig:heartpole} illustrate the performance of our proposed Observation-Constrained MDP (OCMDP) framework across two tasks: the \textit{Diagnostic Chain} Task and the \textit{HeartPole} Task. The results demonstrate significant improvements in both control performance and observation efficiency, validating the versatility and effectiveness of OCMDP across different complex environments.
In the \textit{Diagnostic Chain Task} (Figure~\ref{fig:diagnostic_chain}), our method achieves a relative improvement of \textbf{71\%} in expected cumulative reward over the baseline model-free control policies. This improvement reflects the effectiveness of our control policy optimization in maximizing rewards while accounting for observation costs. Furthermore, the optimization of the observation policy reduces observation costs by \textbf{50\%} compared to fixed and continuous observation strategies, as shown in Figure~\ref{fig:diagnostic_chain}.(b), indicating that OCMDP successfully balances the trade-off between acquiring valuable information and minimizing associated costs. 

In the \textit{Heartpole} Task (Figure~\ref{fig:heartpole}), OCMDP outperforms several baseline reinforcement learning algorithms, with a relative improvement of approximately \textbf{75\%} in episodic return over the next best-performing algorithm, PPO. This improvement highlights OCMDP's ability to effectively balance observation costs with control rewards, particularly in diagnostic and decision-making tasks with high complexity. Figure~\ref{fig:heartpole}.(b) compares different observation strategies within OCMDP, including \textit{optimal observation} (our adaptive strategy), \textit{always observe}, and \textit{never observe}. The optimal observation strategy yields the highest episodic return, achieving approximately \textbf{80\%} higher return compared to the always observe setting, while incurring fewer observation costs. Additionally, it outperforms the never observe setting by a relative improvement of \textbf{90\%} in episodic return, underscoring the importance of selective observation to enhance control performance on the \textit{HeartPole} Task.

Additionally, in the \textit{Diagnostic Chain} task, the convergence behavior observed in the figure shows that OCMDP rapidly attains superior performance within 150k timesteps, while baseline methods like PPO and DDPG require over 200k timesteps to reach their final (lower) performance levels. In the \textit{Heartpole} task, OCMDP achieves stable performance at around 60k timesteps, compared to 80k-100k timesteps needed by baseline algorithms, demonstrating that our iterative optimization process not only improves final performance but also enhances learning efficiency. The coordinated updates between the control and observation policies lead to a locally optimal joint policy, further validating OCMDP's design.

Overall, these results validate the robustness of OCMDP's adaptive observation policy across different task environments. By judiciously selecting the most informative observations, OCMDP maximizes episodic return while minimizing unnecessary observation costs. This capability to balance control performance and observation efficiency underscores OCMDP's potential for applications where cost-efficient information acquisition is essential.



\section{Conclusion and Future Work}

In conclusion, we have introduced a novel approach that effectively balances the costs of information acquisition with the benefits of informed decision-making in cost-sensitive environments. By defining the Observation-Constrained Markov Decision Process and implementing a decomposed deep reinforcement learning algorithm, our approach efficiently manages the expanded action space inherent in simultaneous sensing and control tasks. The successful application of our method to both simulated diagnostic scenarios and the HeartPole healthcare environment highlights its capability to significantly reduce observation costs while maintaining high levels of control performance. These results demonstrate the practicality and superiority of our strategy over existing baseline methods, paving the way for more resource-efficient decision-making systems in real-world applications where the judicious use of observations is paramount.

While our approach shows promising results, there are several avenues for future research to further enhance its capabilities. One potential direction is to extend the Observation-Constrained MDP framework to incorporate multi-agent systems, allowing for collaborative observation and control strategies across distributed agents. This could be particularly beneficial in environments where multiple entities need to make coordinated decisions with shared information costs. Another area of future work is to explore adaptive observation cost functions that dynamically adjust based on task complexity or environmental uncertainty. This would enable the model to prioritize critical observations in more complex or volatile scenarios, further optimizing resource use without compromising decision quality. Additionally, integrating our framework with real-world sensory data and testing in practical applications, such as healthcare monitoring, autonomous vehicles, and industrial IoT systems, could provide valuable insights into its performance under more diverse and realistic conditions. These efforts could pave the way for the development of robust, cost-effective decision-making systems that are tailored to operate effectively in dynamic, resource-constrained environments.

\bibliography{neurips_2022}
\bibliographystyle{named}



\end{document}